\documentclass[sigconf]{acmart}
\AtBeginDocument{%
  }

\setcopyright{acmlicensed}
\copyrightyear{2026}
\acmYear{2026}
\setcopyright{cc}
\setcctype{by}
\acmConference[IH '26]{Interactive Health Conference}{July 05--08, 2026}{Porto, Portugal}
\acmBooktitle{Interactive Health Conference (IH '26), July 05--08, 2026, Porto, Portugal}
\acmDOI{10.1145/3786579.3804996}
\acmISBN{979-8-4007-2422-0/2026/07}

\usepackage{xspace}
\usepackage{subcaption}
\usepackage{wrapfig}
\usepackage{graphicx}
\usepackage{booktabs}
\usepackage{caption}   %
\usepackage{tabularx}  %
\usepackage{array}     %

\newcommand{\pname}{\texttt{PATHFinder Agent}\xspace}

\newcommand{\q}[1]{``#1''}
\usepackage[most]{tcolorbox}

\newtcolorbox{systempromptbox}[1]{ %
  colback=gray!10,
  colframe=gray!80,
  sharp corners,
  boxrule=0.8pt,
  left=6pt,right=6pt,top=6pt,bottom=6pt,
  title={#1}
}

\newcommand{\formchatinterface}{\textbf{Form+Chat Interface}}

\begin{document}

\title{PATHFinder Agent for Tailored Prenatal Care}

\author{Vaibhav Balloli}
\affiliation{%
  \institution{University of Michigan}
  \city{Ann Arbor}
  \country{USA}}
\email{vballoli@umich.edu}

\author{Carissa Samuel}
\affiliation{%
  \institution{University of Michigan}
  \city{Ann Arbor}
  \country{USA}}

\author{Samia Abdelnabi}
\affiliation{%
  \institution{University of Michigan}
  \city{Ann Arbor}
  \country{USA}
}

\author{Alex Peahl}
\affiliation{%
 \institution{University of Michigan}
 \city{Ann Arbor}
 \country{USA}}

\author{Elizabeth Bondi-Kelly}
\affiliation{%
  \institution{University of Michigan}
  \city{Ann Arbor}
  \country{USA}}
\email{ecbk@umich.edu}

\renewcommand{\shortauthors}{Balloli et al.}

\begin{abstract}
  Prenatal care is an important preventive service designed to improve outcomes for pregnant individuals. The American College of Obstetricians and Gynecologists (ACOG) recently introduced guidelines advocating tailored prenatal care, called PATH (Plan for Tailored Healthcare). We present \texttt{PATHFinder Agent} (Planner for Appropriate Tailored Healthcare), an end-to-end conversational agentic system that gathers patient health and social context through structured dialogue, curates individualized prenatal care plans aligned with PATH guidelines, and surfaces community resources from Michigan 211. The system features a four-stage workflow spanning patient intake, dynamic interaction, plan synthesis, and clinician oversight. We evaluate frontier large language models (LLMs) on expert-curated rubrics across five clinical dimensions, finding that GPT-5.2 achieves the highest average score (77.6\%) while identifying key gaps in antenatal testing recommendations. We discuss future validation through human participant studies and randomized controlled trials.

\end{abstract}



\maketitle

\section{Introduction}

Prenatal care is a crucial preventive service that improves pregnancy outcomes for mothers and their children \citep{peahl2020prenatal}, with nearly four million pregnant patients each year in the United States receiving prenatal care. Prenatal care is multifaceted, involving planning and providing medical care, screening tests, answering questions, and connecting people to appropriate social and community resources. 

The American College of Obstetricians and Gynecologists (ACOG), a professional association of physicians specializing in obstetrics and gynecology in the United States, recently proposed a transformation to existing guidelines, with the aim to begin \textbf{``carefully tailoring prenatal care''} \cite{peahl2025tailored}. These guidelines, called PATH (Plan for Appropriate Tailored Healthcare), broadly tackle \textbf{(a) addressing unmet social needs} and \textbf{(b) incorporating alternative care modalities} to help tailor the plan to the patient and provide prenatal care to many more birthing people. PATH was carefully designed after conducting interviews with 110 patients, clinicians, and policy makers representing 25 organizations and more than 75 clinics. 

\begin{figure*}
  \centering
  \begin{subfigure}[b]{0.48\textwidth}
    \centering
    \includegraphics[width=\linewidth]{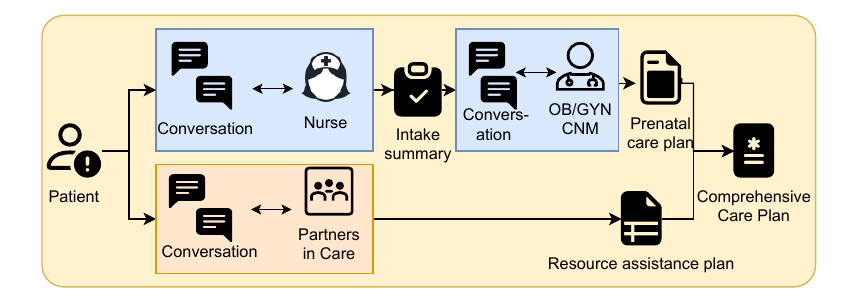}
    \caption{Current practices involve complex, disjoint conversations with patients to provide prenatal care.}
    \label{fig:prenatal_problem}
  \end{subfigure}\hfill
  \begin{subfigure}[b]{0.48\textwidth}
    \centering
    \includegraphics[width=0.9\linewidth]{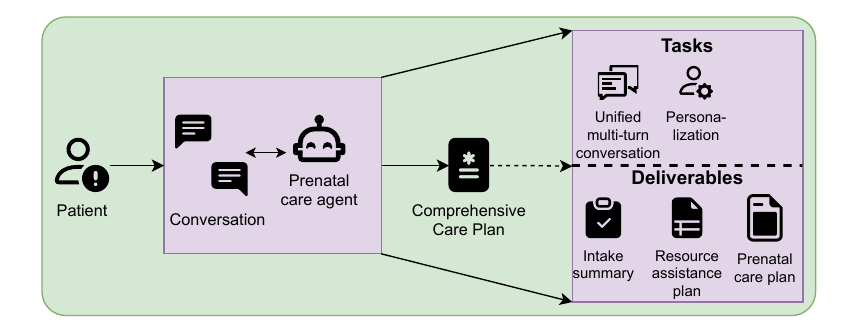}
    \caption{Centralized, scalable conversation with \pname equipped with the latest guidelines and grounded resources.}
    \label{fig:prenatal_proposal}
  \end{subfigure}
  \caption{Illustrations of current practices (left) and our proposed solution (right)}
  \label{fig:both}
\end{figure*}

Adopting PATH is a complex task for both the patients and clinicians (Figure \ref{fig:prenatal_problem}). 
For patients with unmet social needs, additional prenatal visits with a maternity care professional are unlikely to address underlying needs and may create additional burden \citep{peahl2021michigan}. Furthermore, increased demands on physicians and other health care professionals to address patients' unmet social needs with insufficient resources have been associated with burnout \citep{tabata2024physician}.

We %
propose to help mitigate these challenges by creating a conversational AI agent-based interface, \pname, with abilities to \textbf{(a) follow up} with questions and clarifications requiring medical and conversational knowledge, \textbf{(b)} \textbf{perform actions via tool-calls} (searching for resources, generating a report, etc.), and \textbf{(c)} \textbf{follow instructions} to provide a comprehensive plan. We further carefully work towards avoiding unintended behaviors, ensuring oversight and grounding, and testing for deployment, as state-of-the-art medical systems like g-AMIE have called for %
\cite{vedadi2025towards}. We anticipate \pname has the potential to support broad deployment of the PATH guidelines, thereby improving health and well-being for pregnant and birthing individuals.

\section{Background}

We identify three key areas from the literature that have particularly helped us shape the system.

\paragraph{Prenatal Care}
Over 80\% of adverse pregnancy outcomes are preventable through essential prenatal care and the management of unmet social needs \cite{trost2022pregnancy}. However, the traditional 12–14 visit in-person model \cite{kilpatrick2017guidelines} is often inaccessible due to social drivers of health—such as housing instability and inflexible work schedules \cite{semega2021income, nasem2019integrating}—leading many patients to be underserved and even feel unheard  \cite{bellerose2022systematic,betron2018expanding,mohamoud2023vital}. In response, ACOG developed the Plan for Appropriate Tailored Healthcare in pregnancy (PATH) \cite{peahl2021michigan}, which utilizes shared decision-making to tailor care through flexible visit frequencies \cite{peahl2020right, turrentine2023prenatal, balk2023reduced} and telehealth modalities \cite{balk2023televisits}, for example. 
Despite its potential,  gaps remain in implementing PATH effectively within complex healthcare systems \cite{nijagal2021using}, and taking patient preferences into account will be vital \cite{peahl2021patient, peahl2020preferences}.

\paragraph{Medical Agents} Gemini, AMIE~\citep{saab2025advancing,palepu2025towards,vedadi2025towards,johri2024craft} and MedAgentBench~\citep{jiang2025medagentbench} focus on electronic health records, and MedAgentGym focuses on programming-centric agentic tasks in medicine\citep{xu2025medagentgym}. These systems do not focus on integrating both domain-specific needs, like prenatal care, and domain-agnostic needs, like safety measures.

\paragraph{Tool-use and Conversational Systems} To utilize such medical agents to support patients and providers implementing PATH, we build on work where agents must ask clarifying questions and follow instructions to achieve the user's goal, as well as work in which agents need to call external tools.  Research works like \citep{qin2023toolllm,patil2023gorilla} investigated LLMs' capabilities to invoke specific \q{tools} to solve tasks in mathematical reasoning, program synthesis, and general tasks. LLMs can decide to invoke a tool by generating tokens in an expected format, which enables them to navigate autonomously by recursively invoking tools. Instruction-tuned LLMs paired with strategies like ReACT (Reasoning and Acting) \citep{yao2022react} have demonstrated near-perfect capabilities to accurately invoke tools. On the other hand, researchers have also studied task-oriented conversations \citep{budzianowski2018multiwoz,chen2021action} between humans and automated evaluation of conversational models \citep{gur2018taskdialogues}. Subsequent works \citep{yao2024tau,lu2025toolsandbox,barres2025tau2,patil2025bfcl} take a step closer to real-world applications with multi-turn interactions between the human and agents.

\section{\pname: System Design}

\begin{figure*}[t]
    \centering
    \includegraphics[width=0.6\linewidth]{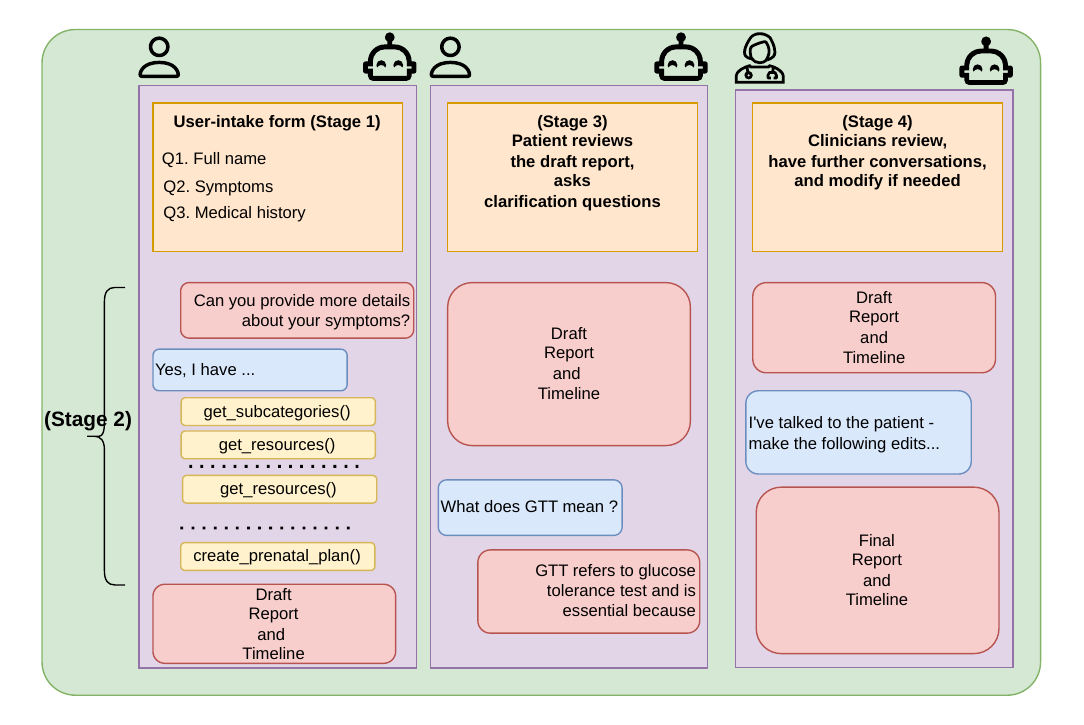}
    \caption{\textbf{PATHFinder interfaces.} \pname has a four-stage overflow: 1) Patient information intake (standardized form), 2) LLM-generated UI for structured, open-ended dialogue, 3) Draft report review and clarification for the patient and 4) Clinician review during provision of care.}
    \label{fig:conversation_breakdown}
\end{figure*}

\subsection{Problem Statement}

\begin{figure}
    \centering
    \includegraphics[width=\linewidth]{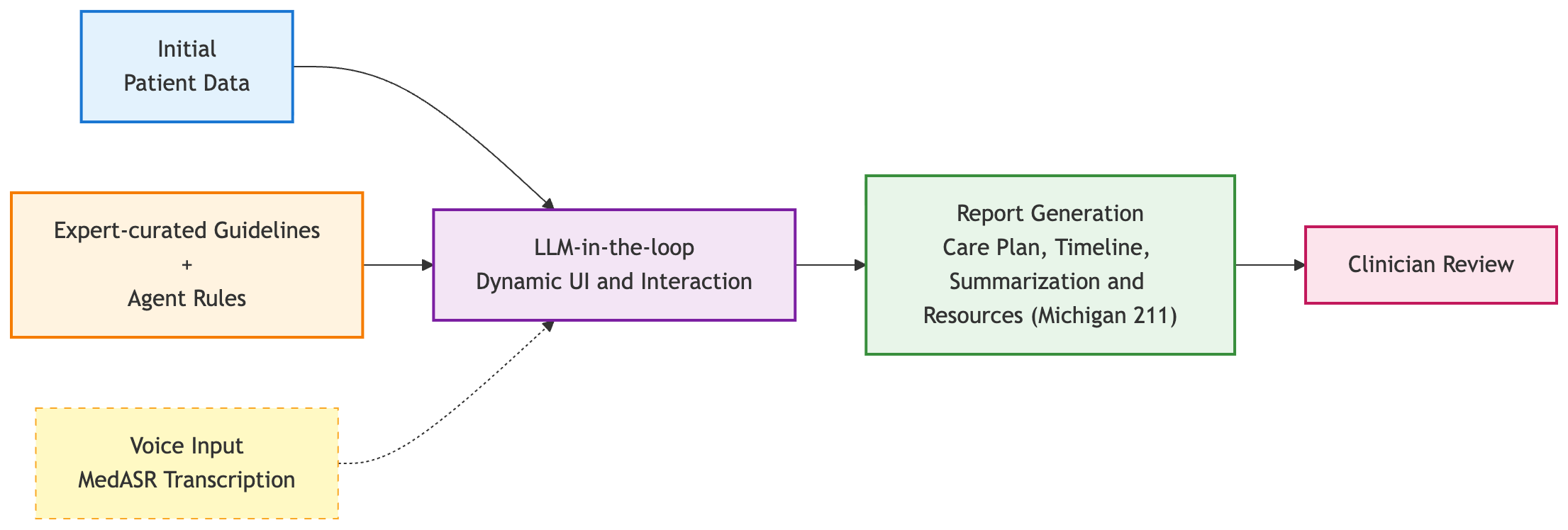}
    \caption{\textbf{System architecture.} React frontend communicates with a FastAPI backend routing to the \pname agent (LLM + tool executor), oversight classifier, FHIR service, and report generator. Conversation state is persisted in a relational database.}
    \label{fig:architecture}
\end{figure}

Given a patient's medical characteristics and social context, \pname must (i) gather relevant information through structured, open-ended dialogue, (ii) curate an individualized prenatal care plan aligned with the specified guidelines, and (iii) surface Michigan 211 community resources matched to identified social needs—while enabling ongoing clinician review and oversight (Figure \ref{fig:architecture} illustrates the workflow).

\subsection{Objectives and Requirements}

Core design requirements include interfaces and question flows that mirror existing clinical intake processes, and targeted, non-redundant follow-up questions with an interaction mode to prevent patient fatigue in long sessions while capturing data efficiently.

\subsection{Design}

\paragraph{Team.} \texttt{PATHFinder} was co-designed by a board-certified Obstetrician and Gynecologist and Certified Nurse Midwife (CNM) alongside two computer scientists, ensuring jointly validated clinical and technical requirements.

\begin{table}
\begin{tabular}{@{}ll@{}}
\toprule
\textbf{Category} & \textbf{Tools} \\
\midrule
Medical           & TOLAC Calculator; Defer to Clinician \\
\midrule
Social needs resources          & Get Groups; Get Categories; \\
                  & Get Subcategories; Get Resources \\
                  & by Subcategory \& ZIP; No Resources Message \\
\midrule
Personalization   & Consult LLM for follow-up questions \\
\midrule
Report            & Add Patient Summary; Add Clinical Summary; \\
                  & Add Recommendations; Add Resources; \\
                  & Create Visit Schedule \\
\bottomrule
\end{tabular}
\caption{Agent tool suite (13 tools across 4 categories).}
\label{tab:tools}
\end{table}

\paragraph{Tools and Orchestration.} The agent is equipped with 13 tools across four categories (Table\ref{tab:tools}). Medical tools include a TOLAC (trial of labor after cesarean) calculator and a structured referral-to-clinician action. Resource tools implement a hierarchical Michigan 211 query interface. A personalization tool consults a secondary LLM to generate context-aware follow-up questions. Report tools incrementally build the patient and clinician summaries. The system instructions consists of $\sim$14,000 tokens, with domain knowledge (from ACOG guidelines\cite{peahl2025tailored}), tool use instructions, workflow and safety policies, where the agent retrieves the knowledge and instructions to orchestrate the conversation.

\paragraph{Social needs resources data.} We use Michigan 211\footnote{\url{https://mi211.org/}} data organized by category (food, housing, transportation, utilities, clothing) and can be queried by subcategory and ZIP code via dedicated tool calls.

\begin{figure*}
    \centering
    \begin{subfigure}[b]{0.48\linewidth}
        \centering
        \includegraphics[width=\linewidth]{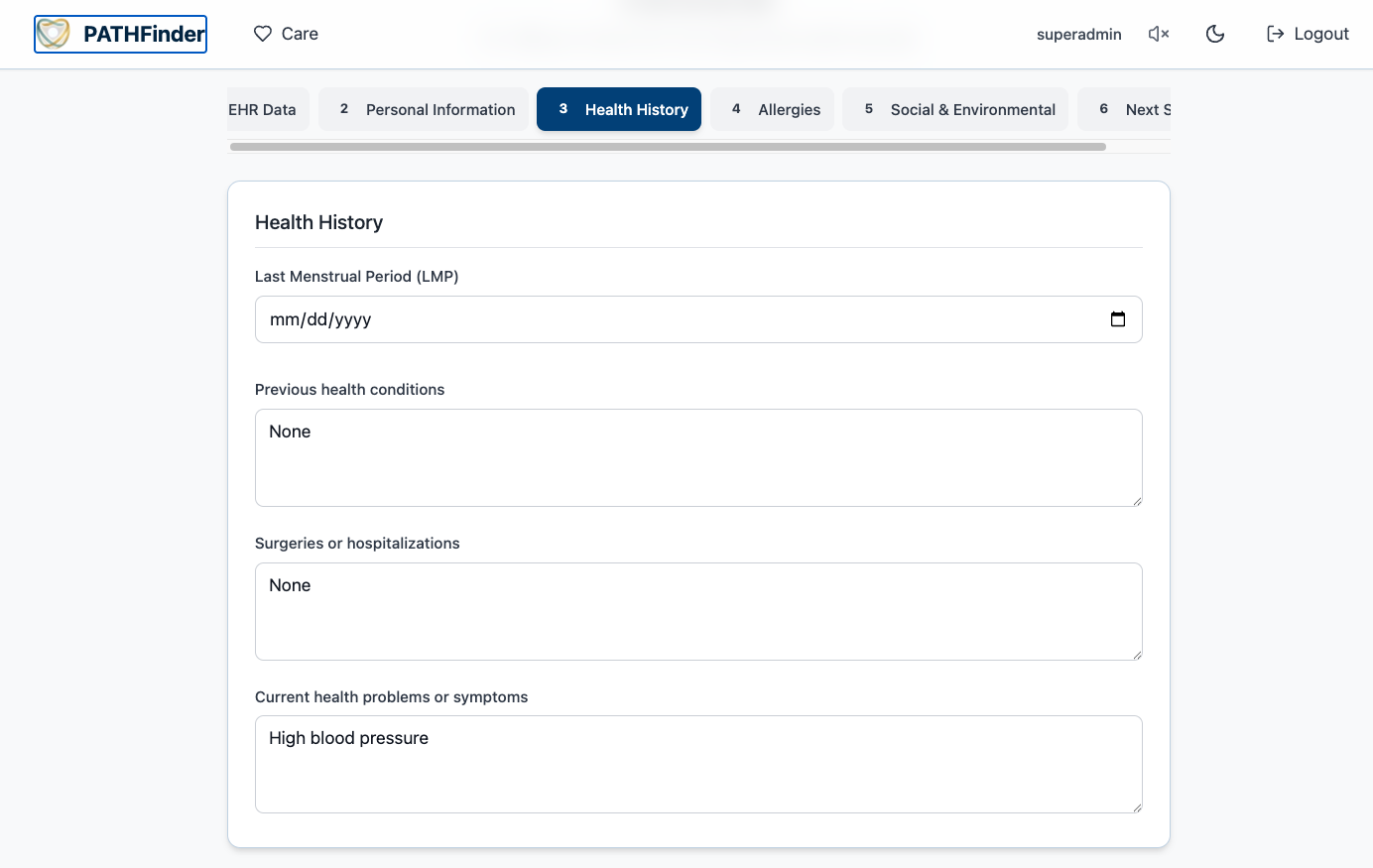}
        \caption{Patient intake form (Stage1).}
        \label{fig:intake}
    \end{subfigure}\hfill
    \begin{subfigure}[b]{0.48\linewidth}
        \centering
        \includegraphics[width=\linewidth]{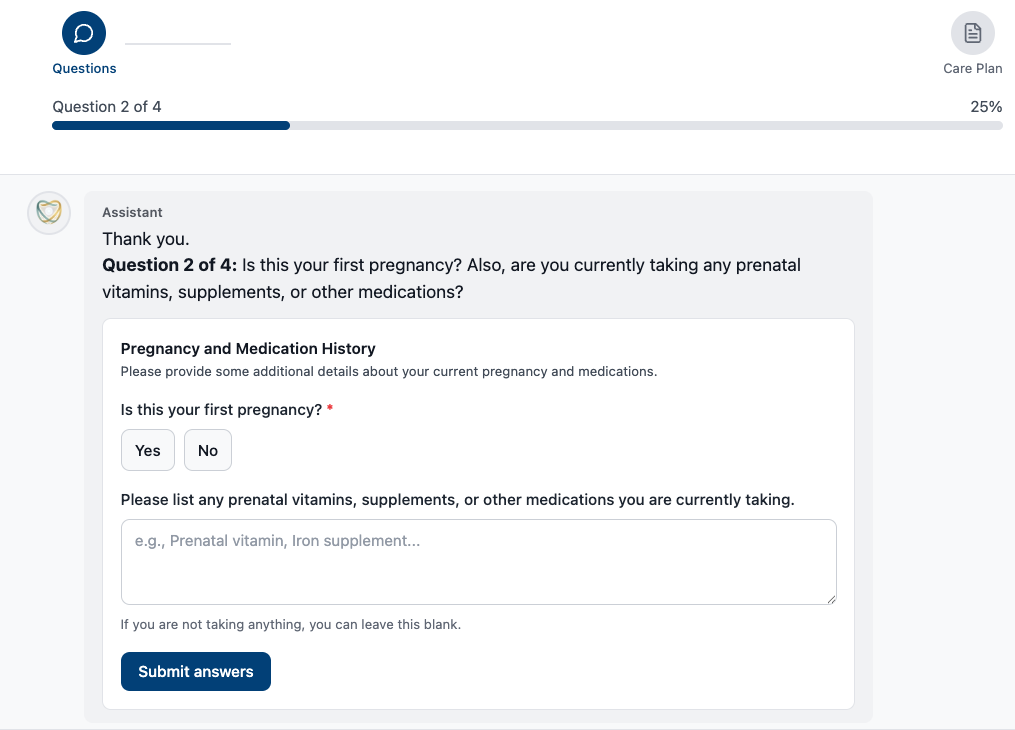}
        \caption{Dynamic interaction with LLM generating intermediate UI.}
        \label{fig:dynamic_interaction}
    \end{subfigure}
    \begin{subfigure}[b]{0.48\linewidth}
        \centering
        \includegraphics[width=\linewidth]{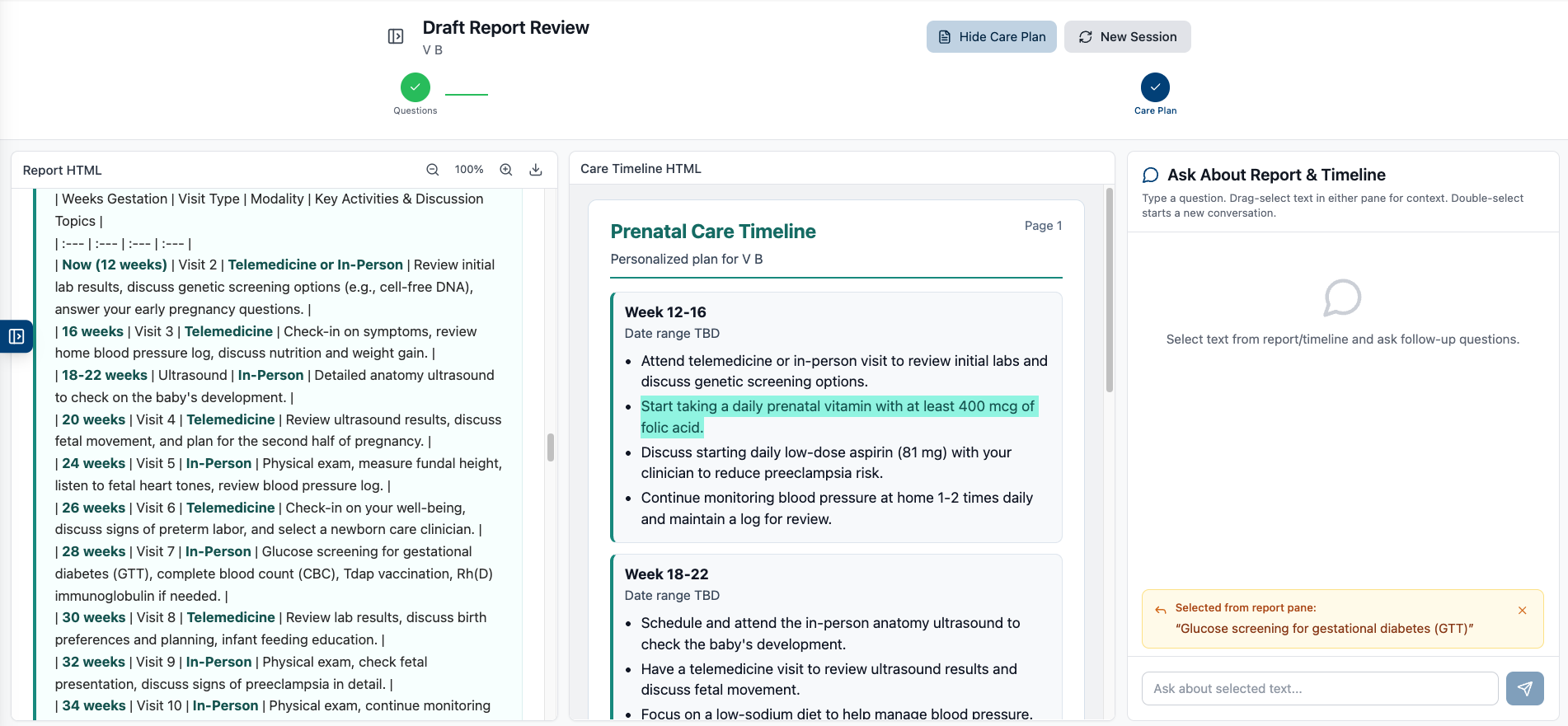}
        \caption{Patient draft report review and clarification (Stage 3).}
        \label{fig:draft_report}
    \end{subfigure}\hfill
    \begin{subfigure}[b]{0.48\linewidth}
        \centering
        \includegraphics[width=\linewidth]{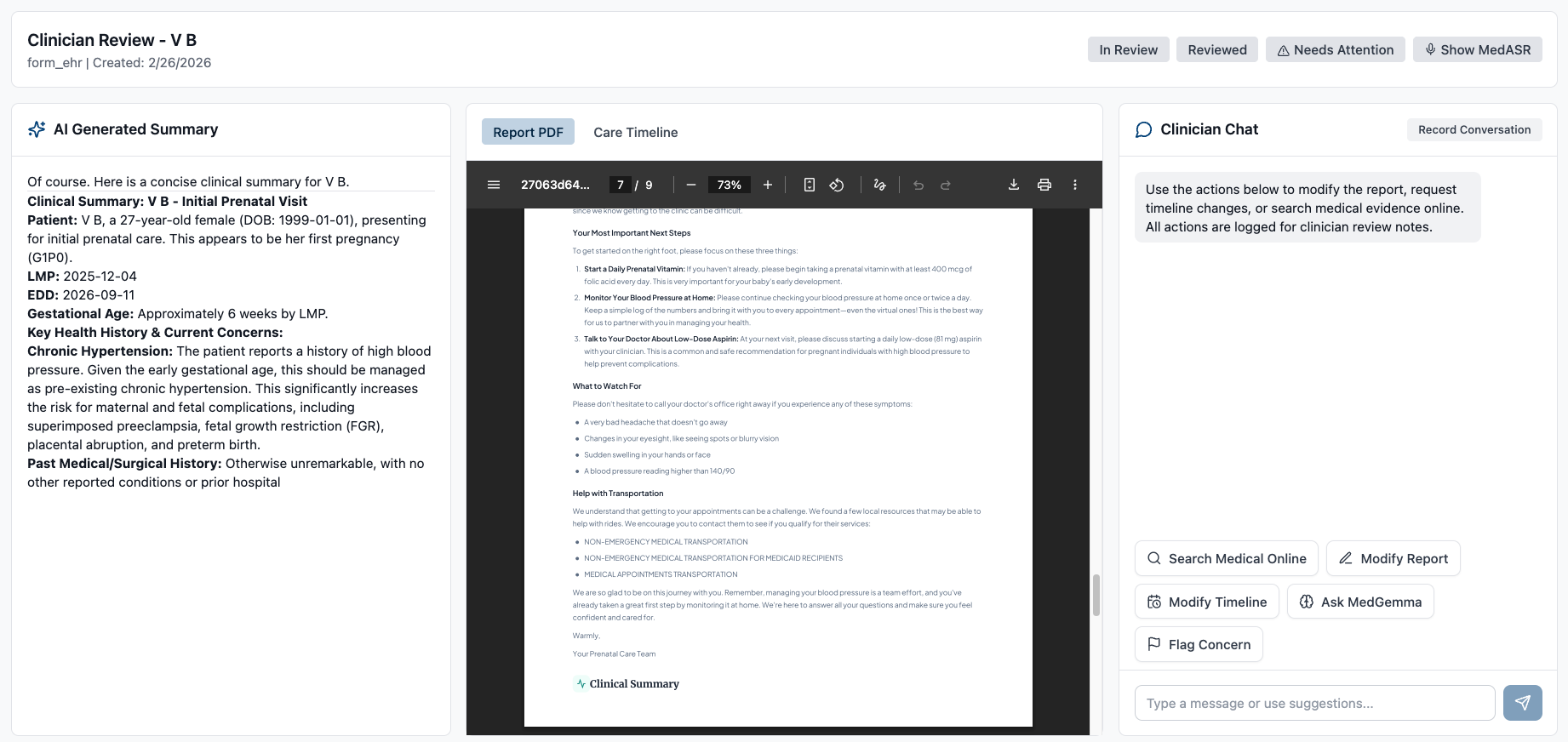}
        \caption{Clinician oversight and review (Stage4).}
        \label{fig:clinician}
    \end{subfigure}
    \caption{\textbf{\pname interfaces.} (a)Structured patient intake form; (b) \pname interviews the user to understand additional medical history details, and social determinants of health factors; (c) Patient reviews the draft report for transparency and clarification; (d)Clinician oversight dashboard showing conversation stage, risk flags, and escalation controls.}
    \label{fig:ui}
\end{figure*}

\paragraph{Interface design.} \pname supports interaction via \formchatinterface, where the patient first completes a standardized intake form mirroring current clinical processes; the agent then transitions to an adaptive conversational phase to ask individualized follow-up questions, where another LLM reviews the questions and generates a form-based interface for the user to interact which supports buttons, check boxes, and switches to reduce the text entered by the user (see Figure \ref{fig:dynamic_interaction}). Based on all the available user information and guidelines~\cite{peahl2025tailored}, appropriate reports and timelines are generated for the clinicians to review and patients to look at and clarify the content. Similarly, the agent is available for the clinician to make edits to the report if needed.

\paragraph{Workflow.} The end-to-end workflow has four stages. \textit{Stage 1 (Intake):} the patient fills a standardized form capturing demographics, medical history, gestational age, and social factors (Figure \ref{fig:intake}). \textit{Stage 2 (Dynamic interaction):} conditioned on responses and guidelines, the agent asks personalized follow-up questions to elicit unmet social needs, preferences, and barriers and uses tools in Table \ref{tab:tools} to compose a plan (Figure \ref{fig:dynamic_interaction}). \textit{Stage 3 (Plan synthesis):} the agent invokes report tools to produce a patient-facing summary, a clinical summary, a visit-schedule recommendation, and curated Michigan 211 resources (Figure \ref{fig:draft_report}).  \textit{Stage 4 (Clinician review):} the clinician reviews the plan with capabilities to approve or edit (Figure \ref{fig:clinician}).

\subsection{Clinician Oversight}

Once the draft plan is generated, the clinician reviews the patient summary, report, and timeline to validate it and change it based on additional conversations and concerns. We highlight that this workflow potentially allows the clinician to better address concerns and provide care to patients while being involved in the prenatal care planning and validation of the plan.

\subsection{Evaluation}

We evaluate \pname on synthetic patient profiles spanning diverse medical histories (e.g., high-risk pregnancies, TOLAC candidates). Each profile pairs with expert-curated rubrics specifying the expectations along the following dimensions: 1) the right visit frequency, 2) the right services (testing, recommendations, etc.), 3) timing for antenatal testing, 4) timing for growth ultrasound, and 5) modality of care to the patient (mandatory in-person, mix of in-person and group health, etc.). 

\begin{figure}
    \centering
    \includegraphics[width=0.8\linewidth]{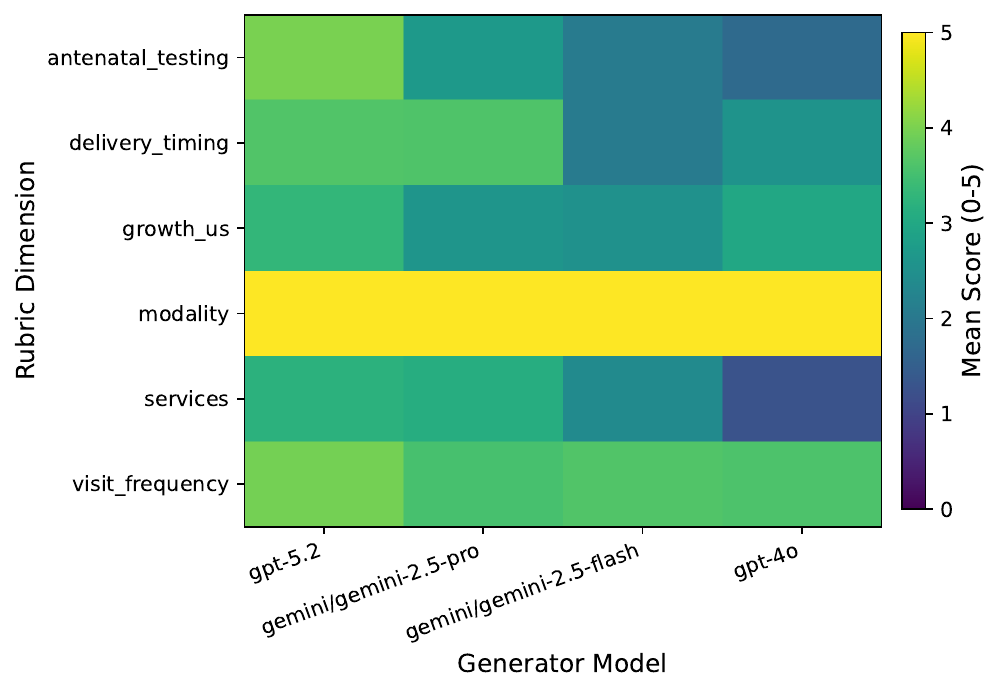}
    \caption{\textbf{Per-dimension rubric scores.} Antenatal testing and services recommendation show the widest performance gap across models (1 is lowest, 5 is highest).}
    \label{fig:heatmap}
    \label{tab:eval}
\end{figure}

We adopt LLM-as-judge scoring to measure the performance of frontier models' recommendations across the five dimensions with the final score per condition normalized to 1. Table \ref{tab:eval} shows aggregate rubric scores across state-of-the-art LLMs from OpenAI (GPT-5.2, GPT-4o) and Google (Gemini 2.5 pro and flash). Furthermore, Figure \ref{fig:heatmap} breaks down scores by dimension, with visit frequency being the easiest across all models and recommending antenatal testing and other services being the most difficult. These results suggest that robust oversight measures, both LLM- and human-driven, must be integrated into these systems with appropriate communication to ensure that deployed models do not make mistakes.

\begin{table}
    \centering
    \begin{tabular*}{\linewidth}{@{\extracolsep{\fill}}lr@{}}
      \toprule
      \textbf{Model} & \textbf{Avg. Rubric Score(\%)} \\
      \midrule
      GPT-5.2            & 77.60 \\
      Gemini 2.5 pro   & 71.50 \\
      Gemini 2.5 flash & 62.00 \\
      GPT-4o           & 57.25 \\
      \bottomrule
    \end{tabular*}
    \caption{Average rubric-based scores (0-100\%) across frontier models. Higher is better.}
\end{table}

\section{Future Work}

Future research will focus on establishing formal accuracy guarantees for \pname to strengthen the system's technical reliability and performance standards. We will concurrently conduct a series of human participant experiments.

\section*{Acknowledgements}
This work was partially supported by funding from Google and the University of Michigan (including the Raoul Wallenberg Institute, E-Health and Artificial Intelligence, and the Center for Academic Innovation). We also thank David Stutz for helpful discussions during the project.

\bibliographystyle{ACM-Reference-Format}
\bibliography{ref.bib}

\end{document}